\documentclass[conference]{IEEEtran}
\IEEEoverridecommandlockouts
\usepackage{amsmath,amssymb,amsfonts}
\usepackage{graphicx}
\usepackage{textcomp}
\usepackage{xcolor}
\usepackage{color}
\usepackage{bm}
\usepackage{amsmath}
\usepackage{amssymb}
\usepackage{amsthm}
\usepackage{mathrsfs}
\usepackage{enumerate}
\usepackage{multirow}
\usepackage{color}
\usepackage{threeparttable}
\usepackage{subfigure}
\usepackage[square, comma, sort, numbers]{natbib}

\usepackage[breaklinks=true, colorlinks, bookmarks=false]{hyperref}

\usepackage{times}
\usepackage{breakurl}
\usepackage{array}
\usepackage{verbatim}
\usepackage{algorithm}
\usepackage{bbding}
\usepackage{algpseudocode}
\usepackage{lettrine}
\usepackage{tabularx}
\usepackage{booktabs}

\def\BibTeX{{\rm B\kern-.05em{\sc i\kern-.025em b}\kern-.08em
    T\kern-.1667em\lower.7ex\hbox{E}\kern-.125emX}}
\begin{document}

\title{D-YOLO a robust framework for object detection in adverse weather conditions\\ 

}

\author{\IEEEauthorblockN{1\textsuperscript{st} Zihan Chu}
\IEEEauthorblockA{\textit{ Faculty of Mathematical \& Physical Sciences} \\
\textit{University College London}\\
London, United Kindom \\
zihan.chu.22\@ucl.ac.uk}

}

\maketitle

\begin{abstract}


Adverse weather conditions including haze, snow and rain lead to decline in image qualities, which often causes a decline in performance for deep-learning based detection networks. Most existing approaches attempts to rectify hazy images before performing object detection, which increases the complexity of the network and may result in the loss in latent information. To better integrate image restoration and object detection tasks, we designed a double-route network with an attention feature fusion module, taking both hazy and dehazed features into consideration. We also proposed a subnetwork to provide haze-free features to the detection network. Specifically, our D-YOLO improves the performance of the detection network by minimizing the distance between the clear feature extraction subnetwork and detection network. Experiments on RTTS and FoggyCityscapes datasets show that D-YOLO demonstrates better performance compared to the state-of-the-art methods. It is a robust detection framework for bridging the gap between low-level dehazing and high-level detection.
\end{abstract}


\section{Introduction}
Object detection is a critical technology in computer vision that has seen widespread application across various fields including manufacturing, agriculture, healthcare, surveillance and security, traffic control and autonomous vehicles as it serves a dual purpose: categorizing and locating objects within an image. In recent years, there has been a significant advancement in object detection methodologies, particularly with the adoption of deep convolution neural networks(DNN). These approaches have shown remarkable effectiveness and accuracy, substantially boosting the development of related fields. 
However, current mainstream object detection algorithms are mostly benchmarked on normal datasets such as MSCOCO\cite{mscoco}, PASCAL-VOC\cite{B-EVGW10} and Imagenet\cite{deng2009imagenet}.
However, while showing high performance with normal images, these methods usually suffer from a decline in detection performance under adverse weather conditions, especially in case of fog, which is one of he most common situation encountered in real-world  scenarios.
\begin{figure}[t]
    \centering
    \includegraphics[width=1\linewidth]{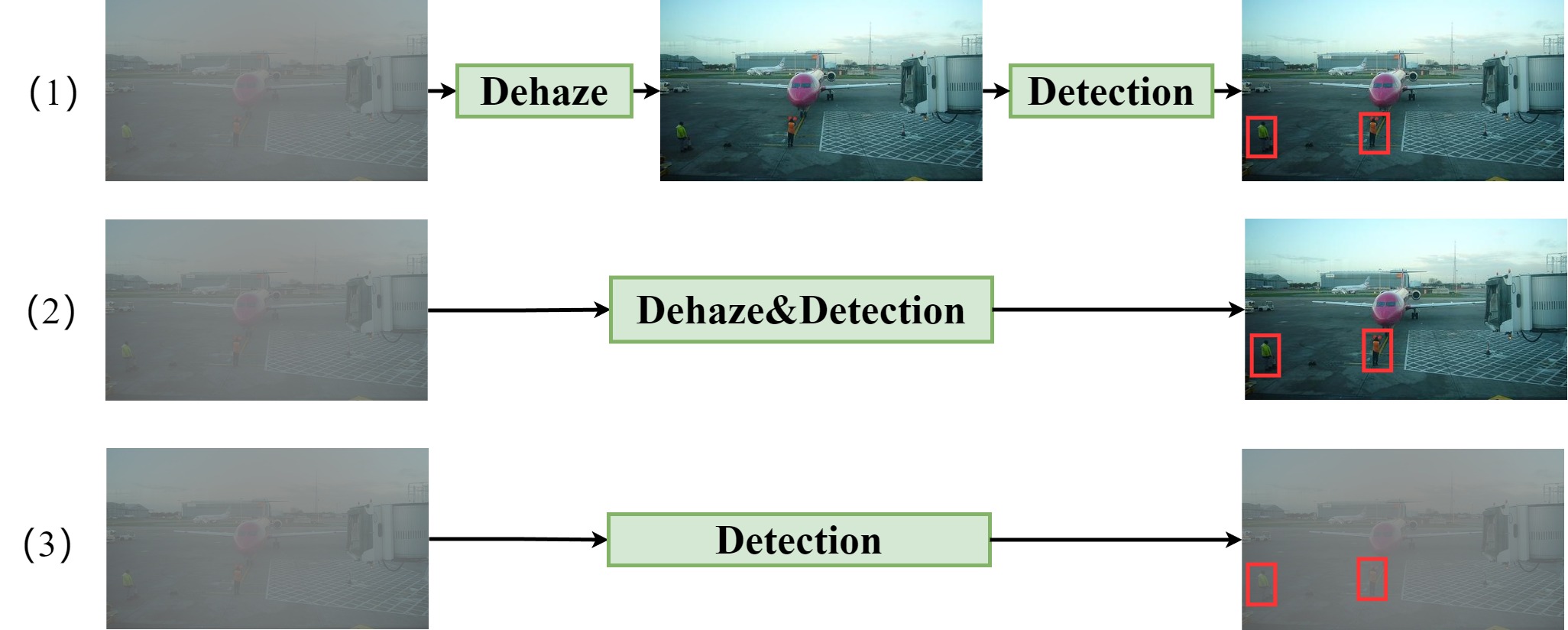}
\caption{Current methods for object detection in adverse weather conditions.(1) Dehazing and detection are performed sequentially. Dehazing models are first trained with synthetic hazy dataset, then the dehazed images are sent to detection networks for object detection. (2) Dehazing and detection tasks are jointly performed in a single network. (3) Detection models are directly trained on hazy dataset.}
\label{intro}
\end{figure}

As shown in Fig \ref{intro}, current solutions can be divided into three classes. The mainstream methods towards object detection in foggy conditions can be categorized into three classes. The most common strategy is to pre-process input images with well developed dehazing algorithms such as AOD-Net\cite{li2017aod}, MSBDN\cite{MSBDN}, Grid-dehazenet\cite{griddehaze}, DCP \cite{he2010single} before feeding them into detection networks. However, these approaches generalize poorly in real world scenarios since the restored images may lose important details. Few works \cite{liu2020connecting} , \cite{qin2020ffa} combine restoration and detection network in a cascade way and optimize the network with a joint loss. Some works use two separate losses for training. But the computational complexity and inference speed are increased, which is unacceptable in resource-constrained scenarios.

To tackle this challenge, we figure it a good way to replace the low-level restoration  with higher level feature adaption. Our D-YOLO comprises three main components, Clear feature extraction subnetwork, Feature adaption subnetwork and Detection subnetwork. The Clear feature extraction subnetwork is responsible for extracting haze free features. Then these resulting features are transferred to the detection subnetwork through feature adaption subnetwork. The . The feature adaption subnetwork yields clean features from the input hazy features. Moreover, to ensure more reliable detection results, we restructure the model into a dual-pathway and design an attention feature fusion module to connect the two branches. This leverages both dehazed and hazy features, further enhancing the model's representation capabilities and improving is performance in complex weather conditions. an All subnetworks and modules in our model are activated in training process. Eventually, the clear feature extraction subnetwork is disabled during inference process for rapid and accurate object detection.

The results from exhaustive experiment on both real-wolrd dataset\cite{B-LRF18} \cite{B-SDVG18}(RTTS, Foggy driving) and synthetic benchmark\cite{cordts2016cityscapes}(FoogyCityscapes) show that our D-YOLO significantly outrun the state-of-the-art object detection approaches. The contribution of our work can be categorized into three aspects.
In summary, there are three main contribution
\begin{itemize}
\item[$\bullet$]We introduce a dual-branch network architecture and an attention feature fusion module, integrating both hazy and dehazed features, resulting in further improvement in detection performance. 

\item[$\bullet$]We proposed an effective and unified way of combining restoration and detection task at feature level, using clear feature extraction subnetwork to provide haze-free information to the detection network. The clear feature extracting module is only activated during training process, resulting in less computational cost at inference process while achieving promising performance. 

\item[$\bullet$]We designed a Feature adaption subnetwork which can transfer haze-free information from the clear feature extraction subnetwork into the detection network to help improve the accuracy of our D-YOLO under adverse weather conditions. 
\end{itemize}



\section{Related Work}\label{sec:rw}

\subsection{Object detection}  
Object detection is one of the most crucial tasks in the field of computer vision, with widespread applications in various scenarios including traffic, medical, remote sensing, etc. With the advancement of computing power and the development of deep learning algorithms, CNN-based deep learning networks have become the mainstream of current object detection works. Object detection methods can be categorized into anchor-based methods including one-stage and two stage, anchor-free methods and transformer-based approaches.

For two-stage methods, the first step is to select region proposals, and then classification and regression are performed on region proposals. As a well-recognized algorithm, RCNN\cite{B-GDDM14} first developed a Region Proposal Network (RPN) for the generation of region proposals, after which feature extraction and detection are performed on selected proposals. The great success of R-CNN comes with a great number of variants based on such framework, such as Fast R-CNN\cite{B-Gir15}, Faster R-CNN\cite{B-Gir15}, Mask R-CNN\cite{he2017mask}, Cascade R-CNN\cite{cai2018cascade}, Libra R-CNN\cite{B-PCS19} and Dynamic R-CNN\cite{B-ZCM20}. Compared with single-stage method, in spite of its significance in accuracy, region-proposal-based methods still suffer from low inference speed, which hinders its application in real-time scenarios.

In single-stage methods region proposals and detection results are generated simultaneously, presenting a faster inference speed and a slight decline in detection performance. YOLO\cite{A-21} series are the most popular single-stage networks. It divides the input image into multiple grids and each grid is assigned with several anchors, covering various shapes and sizes that objects in the image might take. After the great success of YOLOv3\cite{A-23}, multiple improvements and numerous variants\cite{A-22}\cite{A-24}\cite{A-25} have been established. As another typical kind of single-stage method, SSD series\cite{A-26}\cite{A-27}\cite{A-28}\cite{A-29} adopts anchor sets and performs detection on feature maps of different resolutions. 
\subsection{Object detection in adverse weather conditions}  
Most mainstream detection networks are predominantly designed for general scenarios and optimized for high-quality images under normal weather conditions. Consequently, their detection accuracy tends to decline in case of poor lighting or obstructions like fog, making object detection under adverse weather conditions a challenging task yet to be fully addressed. The key point of improving detection performance in adverse weather conditions is to find the optimum way of combining restoration and detection tasks.

Existing methods can be divided into two classes depending on the order of the restoration and detection tasks. The most common methods are to preprocess low quality images with existing restoration algorithms to remove haze\cite{AODnet}\cite{griddehaze}\cite{he2010single}\cite{MSBDN} or rain\cite{B-LQS19}\cite{B-RLHS20a}\cite{B-DWW20}. The preprocessed images are then fed into object detection networks to generate detection results. Despite that these methods do improve the overall quality of the input images, there’s no guarantee that such approach benefits the performance of the detection network, as the preprocessed images may lack important latent features. Liu et al.\cite{B-LRY22} develop image-adaptive yolo, an end to end framework which perform dehazing and detection in a cascade manner. Some works have attempted to conduct image restoration and object detection simultaneously to mitigate the effects of weather-specific information. Huang and colleagues\cite{dsnet} developed a dual-subnet detection framework consisting of a recovery subnet and a detection subnet. The recovery subnet, trained on ImageNet, is tasked with transforming obscured image features into clear ones, thereby mitigating the impact of blurriness caused by foggy conditions. Recently, some methods treat object detection in adverse weather conditions with domain adaption\cite{B-SUHS19}\cite{B-LRY22}\cite{B-HR21}\cite{B-RSA21}. They assume that there's obvious domain shift between clear images used for training and images under adverse weather conditions. Adversarial training is the most widely used strategy within these methods. For example,Sindagi et al.\cite{B-SOYP20} propose a prior-based unsupervised domain adaptive network for detection in adverse weather. 

\begin{figure*}[t]
    \centering
    \includegraphics[width=1\textwidth]{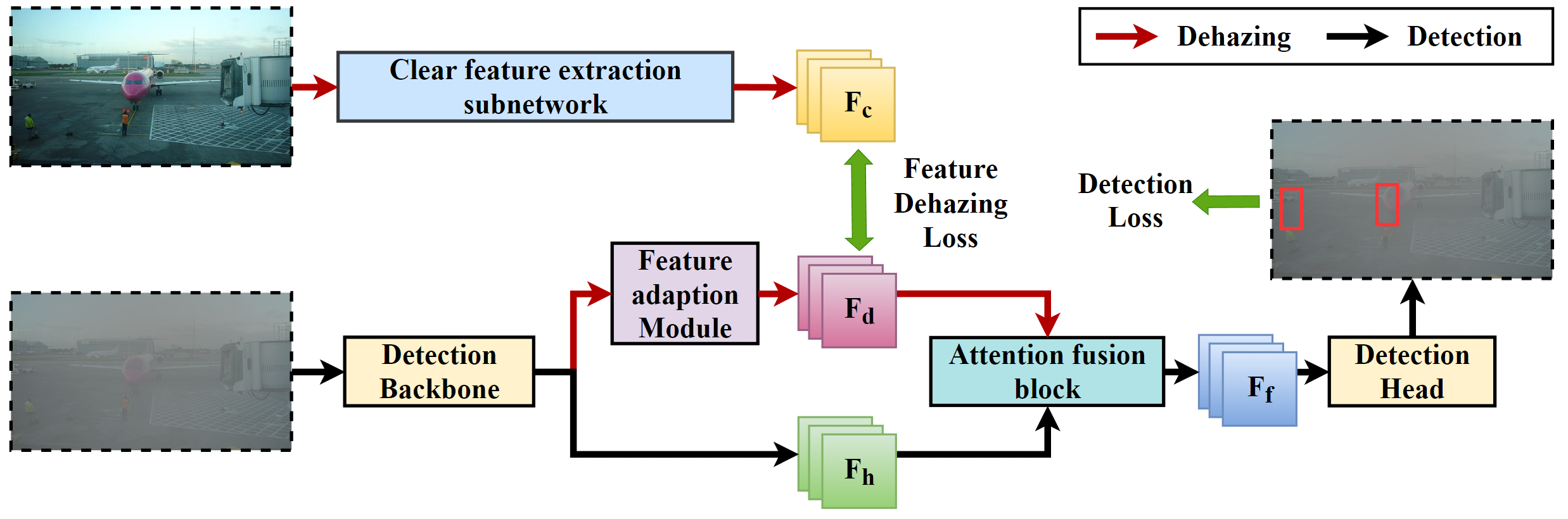}
\caption{The architecture of our proposed D-YOLO. It is composed of a Clear feature extraction subnetwork and a detection network. D-YOLO also adopts a dual-branch structure. One is for yielding dehazed features via feature adaption and the other preserves hazy features. In addition, an attention feature fusion module is introduced to combine different features, after which the fused features are sent to the detection head for bounding box prediction. $F_c$ stands for clear features from the clear feature extraction subnetwork, $F_d$ stands for dehazed features and $F_h$ stands for fused features.  }
\label{net}
\end{figure*}
Zhang et al.\cite{B-ZTHJ21} design a domain adaptive yolo to improve cross-domain performance of one-stage models.

\section{Methodology}\label{sec:method} 
\subsection{Overview}
The overall archietecture of our proposed D-YOLO is demonstrated in Figure \ref{net}. Unlike most existing methods, we consider dealing with the challenge of object detection in adverse weather from a different angle. Sindagi et al.\cite{B-SOYP20} have proposed that threre is domain shift between the clean images and hazy images. In light of this, we consider coping with this challenge with feature adaption. First, we design a feature adaption module along with a feature extraction subnetwork which jointly help the detection network generate haze-free features from input hazy images. Besides, in order to preserve important information from hazy features, we also design an attention feature fusion module to combine dehazed and original hazy features.

\subsection{Clear feature extraction network}  
In our proposed network, the clear feature extraction subnetwork(CFE) is responsible for exploiting the characteristics of the input clear images and sharing them with the feature adaption module. The features extracted from foggy images contain contaminated information. Thus, clear features from CFE subnetwork offer great potential. These features can significantly aid our model in clearly recognizing objects within images throughout the training phase.

In the past decade, many deep convolutional networks have been developed to increase the ability of exoloiting characteristics from images, among which VGG16\cite{VGG}, ResNet50\cite{resnet}, and DarkNet53\cite{redmon2018yolov3} are some of the most seminal. These architectures have been widely adopted as backbone for feature extraction in various computer vision tasks due to their robust performance and architectural innovations. In this study, we have chosen DarkNet53 as the clear feature ectraction subnetwork to extract semantic information from clear images. 
 Multiscale features are selected as the resulting featuremaps($F_c$), these features are then passed to a 1×1 convolutional layer before sending them to the feature adaption subnetwork. Because the channels between $F_c$ and Fa might be different. To be noticed, the CFE subnetwork is only activated during the training phase.
\begin{table}[t]
\centering
\caption{The archietecture of our proposed Feature Adaption module. $C_{in}$ denotes the size of channel dimension of the input features. ODConv represents the omni-dimensional dynamic convolution and CBAM is the convolutional block attention module }
\label{FA}
\begin{tabular}{@{}cccc@{}}
\toprule
Layer     & Kelnelsize & Output Dimension & Stride \\ \midrule
ODConv\_1 & 1×1        & 1×Cin            & 1      \\
ODConv\_2 & 1×1        & 2×Cin            & 1      \\ 
CBAM      & -          & 1×Cin            & -      \\ \bottomrule
\end{tabular}
\end{table} 
\subsection{Feature adaption module}  
The FA module serves as an adapter for learning the profitable information s from clear feature extraction subnetwork. The transferred features are then fed into attention fusion model. The structure of the FA module is shown in table  1. In our proposed D-YOLO, 3 scales of features from the backbone of the detection network are utilized to equip the FA module. 
The structure of the feature adaption module is shown in Table \ref{FA}. As we can see, the feature adaption module consists of two convolution layers and convolutional block attention module\cite{woo2018cbam}(CBAM). 
The channel wise KL divergence loss is utilized in the adapter to optimize and stabilize the training process, thereby bridging the gap between $F_c$ and $F_d$.
The loss function used for training the FA module is expressed as follows:

\begin{equation}
\begin{split}
\text{loss} &= \sum_{i=1}^{C} \Bigg[ \frac{1}{N} \sum_{j=1}^{N} \Bigg( \sum_{k=1}^{H \times W} \text{softmax}(F_c^{(j)}[i, k] / \tau) \\
&\quad \times ( \log \text{softmax}(F_c^{(j)}[i, k] / \tau) \\
&- \log \text{softmax}(F_d^{(j)}[i, k] / \tau) ) \Bigg) \Bigg] \times \tau^2
\end{split}
\end{equation}

Where $F_c$ and $F_d$ denotes the feature map from the CFE module and FA module. $\tau$ denotes the distillation temperature, which is set to 1 in the experiments. In addition, to enhance the performance of the FA module, we apply different weight on different scales of features. As shown in Equation, in our experiments,$\lambda_1$, $\lambda_1$, $\lambda_1$ are set to 0.7, 0.2, and 0.1 since low-level feature maps supply valuable knowledge at a high resolution, which enhances object information.
\begin{figure}[t]
    \centering
    \includegraphics[width=1\linewidth]{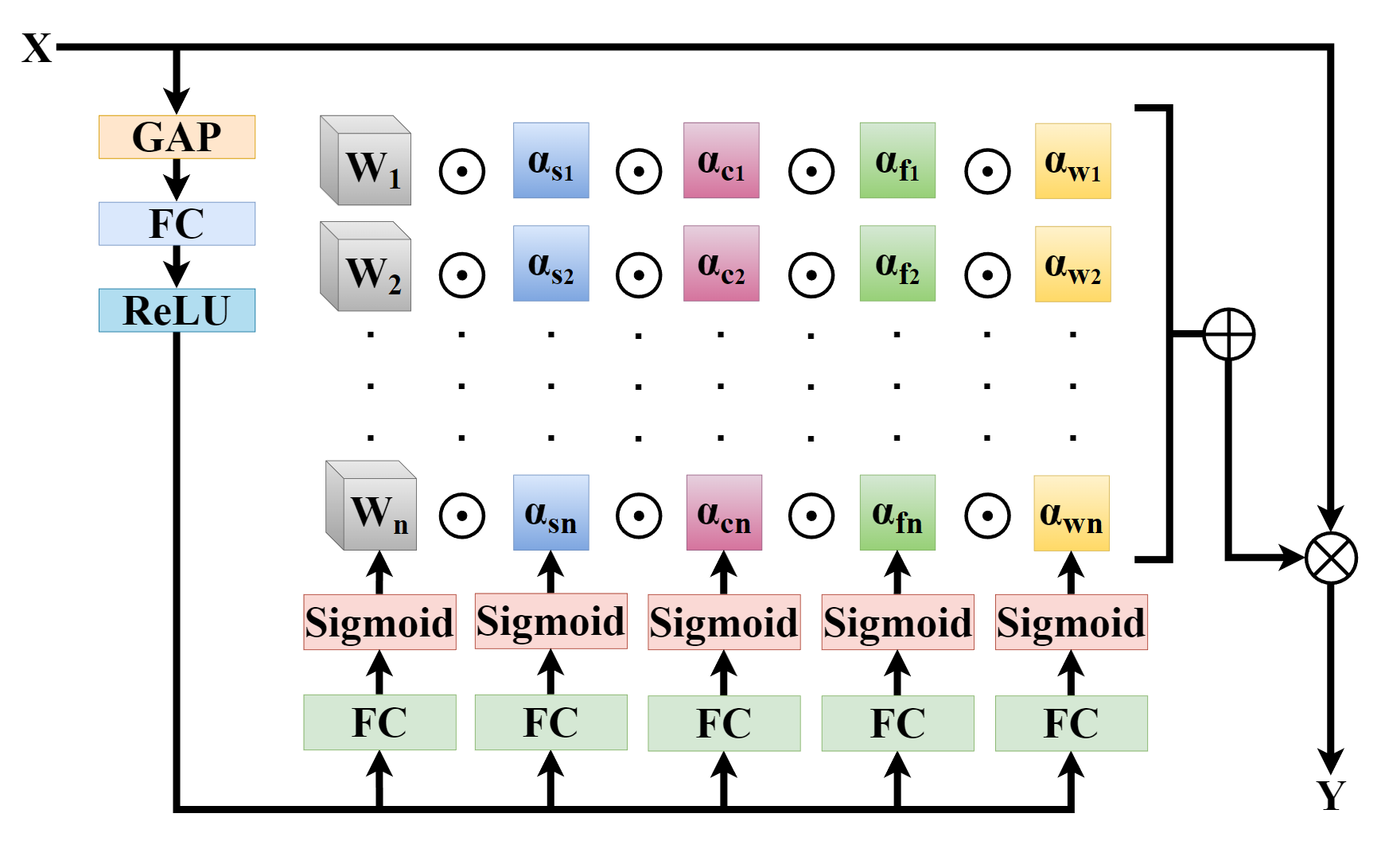}
\caption{The archietecture of Omni-dimensional dynamic convolution. ODConv adopts multi-dimensional attention mechanisms on different convolution, which can build positive dependencies around every element, enhancing the feature extraction, feature transfer ability of the network, resulting in better performance for our feature adaption module. }
\label{fig:odconv}
\end{figure}
\subsection{Omni-dimensional dynamic convolution}
Omni-dimensional dynamic convolution(ODConv), proposed by Li et al.\cite{li2022omni}, is an enhanced version of CNN structure, which adopts multi-dimensional attention mechanisms on different convolution kernels to achieve dynamic covolution. This novel method can greatly improve the feature extraction and feature representation capabilities in all aspects. In our D-YOLO, ODConv is adopted as the convolution layer inside the feature adaption module.

\begin{figure}[t]
    \centering
    \includegraphics[width=1\linewidth]{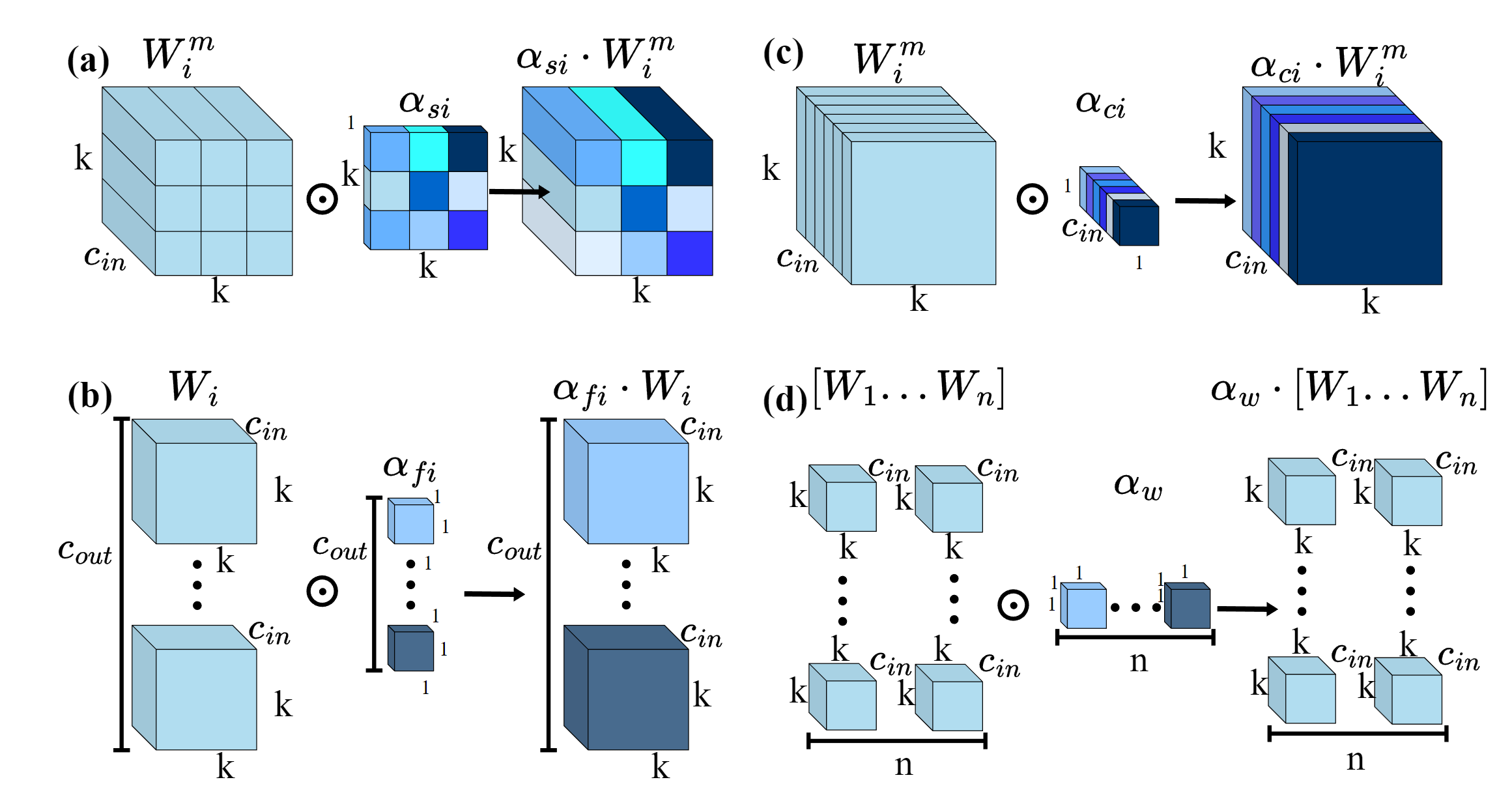}
\caption{Illustration of four types of attentions in ODConv. (a) Location-wise multiplication, (b) channel-wise multiplication, (c) filter-wise multiplication, (d) kernel-wise multiplication.}
\label{fig:odconvattn}
\end{figure}

The architecture of ODConv network is shown in Fig \ref{fig:odconv} Given an input feature map x, odconv first squeeze it into a feature vector with the length of input channel through channel-wise global average pooling(GAP)operations. Then ,the feature vector subsequently pass through a fully connected layer and  four head branches. Each head branch consists of a FC layer and a softmax or sigmoid function, generating normalized attentions $\alpha_{si}$, $\alpha_{ci}$, $\alpha_{fi}$, $\alpha_{wi}$. Fig \ref{fig:odconvattn}

Each attention represents a unique multipilation, including location-wise, channel-wise, filter-wise and kernel-wise. These four types of attentions are then progressively muultiplied to the n convolutional kernels $W_{i}$, giving odconv the ability to consider both spatial and channel information at the same time, thus enhancing the feature representation capability of CNNs. In our method, we adopt ODconv as the convolution layer of the feature adaption module for better adaption and feature extraction ability. In this way, we can acquire more accurate dehazed features, resulting in better detection performance.
\subsection{Attention feature fusion module}
\begin{figure}[t]
    \centering
    \includegraphics[width=1\linewidth]{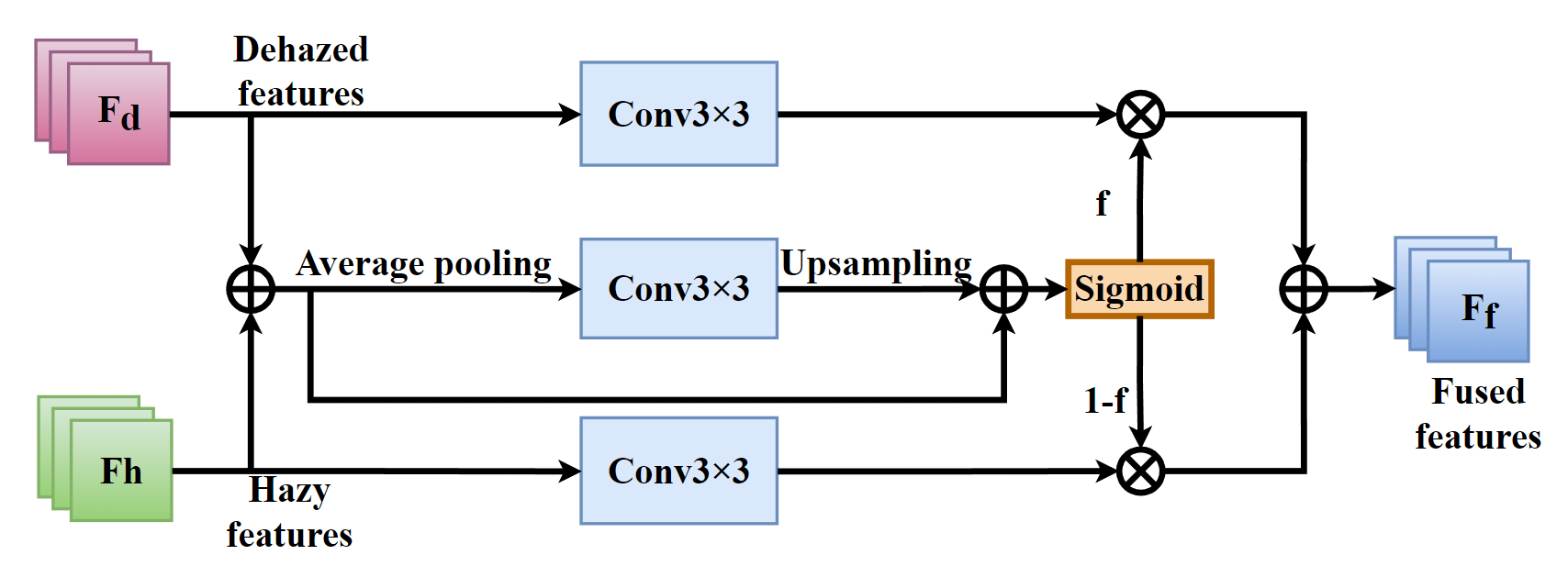}
\caption{Archietecture of our proposed attention feature fusion module. As we can see, in our attention feature fusion module, features are fused through attention calibrated convolutions. }
\label{af}
\end{figure}

In order to better combine hazy and dehazed features, a unique hazy-aware attention feature fusion module(AF) is proposed. Dehazed features may contain contaminated information and may lead to bad performance when the dehazing module performs poorly. Thus, it is essential to build a fusion module to solve the sematic inconsistency between the dehazed features and original hazy features. The archietecture of our attention feature fusion module is shown in Fig \ref{af}

Specifically, first we operate point-wise addition on haze and dehazed features to obtain a fused feature X. Subsequently, we apply $r\times r$ average pooling on the fused feature, enlarging the reception field. The feature is then pass through convolution with kernel $3\times 3$, and a bilinear interpolation operator sequentially. In addition, a shortcut connection is added after the upsampling operation, resulting in T These operations can be denoted as function (\ref{f1})

\begin{equation}
T = X + UP(f_{\text{conv}}(\text{Pool}(X)))
\label{f1}
\end{equation}

Then, we sent the feature T into a sigmoid function, normalizing it into an attention map. Furthermore, we apply convolution with kernel $3\times 3$ on the original input features $F_d$ and $F_h$,  Finally, we obtain the feature $F_f$ by combining the attention map and the convoluted input features through element wise multiplication. The process can be denoted as (\ref{f2})

\begin{equation}
F_f = f_{\text{conv1}}(F_d) \cdot \text{Sig}(T) + f_{\text{conv2}}(F_h) \cdot (1 - \text{Sig}(T))
\label{f2}
\end{equation}
Different from existing attention fusion methods which depends on the channel dimension, our module acquires r×r average pooling instead of global average pooling.Each spatial location is allowed to not only adaptively consider its surrounding informative context as embeddings from the original scale space, but also model inter-channel dependencies. Thus the fields-of-view for the convolution layer is significantly enlarged without increasing too much computation complexity.Second, AFF encodes both hazy and dehazed features. It can preserve important spatial information which is crucial for generating discriminative and selective attention maps for target locations.
\section{Experiment and Analysis}\label{sec:exp}  

\subsection{Dataset}  

We trained the proposed network by the following optimization scheme: For the dataset, due to the fact that there are limited available datasets for object detection in adverse weather conditions, a foggy detection dataset based on the VOC dataset is established. In order to obtain hazy image, we employ the well-known atmospheric scattering model to generate synthetic haze, which is demonstrated by the following equation.
\begin{equation}
I(x) = J(x)t(x) + A(1 - t(x))
\end{equation}
Where J(x) denotes the clean image, A refers to the global atmospheric light, and t(x) refers to the medium transmission map. which can be calculated by:
\begin{equation}
t(x) = e^{-\beta d(x)}
\end{equation}
where $\beta$ denotes the atmosphere scattering parameter, and d(x) refers to scene depth, which can be defined as:
\begin{equation}
d(x) = -0.04 \times \rho + \sqrt{\max(w, h)}
\end{equation}
where $\rho$ denotes the Euclidean distance from the current pixel to the central pixel, w and h refer to the numbers of rows and columns of the image. In our experiments, we set the global atmospheric light parameter A to 0.5, while randomly setting the atmospheric scattering parameter $\beta$ between 0.07 and 0.12 to control the fog level. To make the labels in different datasets complient, we only adopt images in VOC dataset that contains object classes of the RTTS dataset (i.e., car, bus, motorcycle, bicycle, and person) to build the trainset. After processing these clean images on the original VOC dataset, we obtain 9578 foggy images for training (VOC-Foggy).
\begin{table}[t]
\centering
\setlength{\tabcolsep}{6pt}
\caption{Details of our adopted datasets. FCS is the FoogyCityscapes dataset. FDD represents the Foggy Driving dataset.}
\label{datasets}
\begin{tabular}{@{}ccccccc@{}}
\toprule
Dataset & Images & Person & Bicycle & Car   & Motor & Bus  \\ \midrule
VOC-FOG & 9578   & 13519  & 836     & 2453  & 801   & 684  \\
RTTS    & 4320   & 11366  & 698     & 25283 & 1232  & 2585 \\
FCS     & 491    & 3954   & 1171    & 4667  & 149   & 98   \\
FDD     & 101    & 269    & 17      & 425   & 9     & 17   \\ \bottomrule
\end{tabular}
\end{table}
\subsubsection{Testset}
Considering that there are only few public real world detection datasets for adverse weather conditions.In order to evaluate and compare the performance of our proposed D-YOLO and other detection methods in adverse weather conditions. We selected 3 different testsets which includes one synthetic fog datasets and two real world foggy datasets.

Foggy Driving Dataset\cite{B-SDVG18} is a real-world foggy dataset which is used for object detection and semantic segmentation. It involves 466 vehicle instances (i.e., car, bus, train, truck, bicycle, and motorcycle) and 269 human instances (i.e., person and rider) that are labeled from 101 real-world foggy images. Furthermore, although there are eight annotated object classes in the Foggy Driving Dataset, we only select the above-mentioned five object classes for detection to ensure consistency between training and testing.

RTTS\cite{B-LRF18} is a relatively comprehensive dataset available in natural foggy conditions, which comprises 4322 real-world foggy images with five annotated object classes. Considering hazy/clean image pairs are difficult or even impossible to capture in the real world, Li et al. proposed the RTTS dataset to evaluate the generalization ability of dehazing algorithms in real-world scenarios from a task-driven perspective.

FoggyCityscapes\cite{cordts2016cityscapes} is a synthetic foggy dataset which simulates fog on real scenes. Each foggy image is rendered with a clear image and depth map from Cityscapes. Thus the annotations and data split in Foggy Cityscapes are inherited from Cityscapes.There are overall 34 classes in the FoggyCityscapes dataset. same as the Foggy Driving dataset, we filter the images and labels according to the above-mentioned five classes.


\subsection{Implementation Details} 
D-YOLO is trained with SGD optimizer with an initial learning rate of 0.01. We also use a Cosine annealing decay strategy to alter the learning rate. We set the number of epochs to 100 and batch size to 16.
During training, apart from foggy image from VOC-Foggy dataset, we also send the original clear image to the CFE subnetwork for  extracting clear features and sharing them with the detection network. Mosaic data augmentation are disabled during the training process as such strategy might increase the difficulty of training the feature adaption subnetwork which affects the overall performance of the entire network. Feature adaption subnetwork and detection network are firstly jointly trained for 30 epochs. In the remaining 70 epochs, the weights within the feature transfer module are frozen, as the loss of

\begin{table*}[t]
\centering
\setlength{\tabcolsep}{12pt} 
\caption{Comparison of D-YOLO with multiple state-of-the-art methods on RTTS dataset. Red represents the best result, and blue represents the second best result. VOC-f represents the VOC-Foggy dataset, and VOC-c represents the normal VOC dataset.}
\label{RTTS}
\begin{tabular}{@{} c c c c c c c c c @{}}
\toprule 
Method            & Type           & Train Dataset & Person & Bicycle & Car   & Motor & Bus   & All   \\
\midrule 
Yolov8\cite{yolov8}   & Baseline       & VOC-f         & 0.623  & 0.387   & 0.465 & 0.273 & \textcolor{blue}{0.161} & \textcolor{blue}{0.381} \\
Yolov8-C\cite{yolov8} & Baseline       & VOC-c         & 0.619  & 0.364   & 0.157 & 0.241 & 0.155 & 0.367 \\
\midrule 
AOD-YOLOv8\cite{AODnet}      & Dehaze\&Detect & VOC-f, VOC-c  & 0.598  & 0.358   & 0.407 & 0.233 & 0.13  & 0.345  \\
MSBDN-Yolov8\cite{MSBDN}      & Dehaze\&Detect & VOC-f, VOC-c  & 0.589  & 0.374   & \textcolor{blue}{0.393} & 0.209 & 0.12  & 0.337 \\
Griddehaze-Yolov8\cite{griddehaze} & Dehaze\&Detect & VOC-f, VOC-c  & 0.612  & 0.386   & 0.453 & 0.258 & 0.146 & 0.371 \\
DCP-Yolov8\cite{he2010single}        & Dehaze\&Detect & VOC-f, VOC-c  & 0.621      & 0.393       & 0.417     & 0.237     & 0.139  & 0.361     \\
\midrule 
IA-Yolo\cite{IA-Yolo}           & Union          & VOC-f         & \textcolor{red}{0.671}  & 0.353   & 0.414 & 0.211 & 0.136 & 0.357 \\
DSNet\cite{dsnet}             & Union          & VOC-f, VOC-c  & 0.566  & 0.345   & 0.402 & 0.198 & 0.124 & 0.327 \\
MS-DAYolo\cite{ms-dayolo}         & Union          & VOC-f, VOC-c  & 0.637  & 0.391   & \textcolor{blue}{0.479} & \textcolor{blue}{0.281} & 0.157 & \textcolor{blue}{0.389} \\
Ours              & Union          & VOC-f, VOC-c  & \textcolor{blue}{0.658}  & \textcolor{red}{0.402}   & \textcolor{red}{0.538} & \textcolor{red}{0.308} & \textcolor{red}{0.242} & \textcolor{red}{0.430}  \\
\bottomrule 
\end{tabular}
\end{table*}
feature adaption task converges more rapidly compared to the detection task, jointly training the whole process might hinder the accuracy of the detection result. We train our model on a single RTX3090GPU.

\subsection{Comparison with State-of-the-art Methods}
In our experiments, mean average precision (mAP) is chosen as the quantitative evaluation metric on the proposed VOC-Foggy dataset. The comparison is conducted among ten different algorithms. We choose YOLOv8n as our baseline. As shown in Table \ref{RTTS}, the chosen algorithms can be categorized into three classes: (1)Baseline: baseline models are directly trained on hazy or clear images. (2) Dehaze\&detect(D\&D): the hazy images are treated with a two stage method, hazy images first pass through the dehazing models, after which are feed into the pre-trained baseline model.  (3)Union: dehazing and detection task are trained simultaneously on hazy images.
\subsubsection{Comparisons on Real-World Dataset}
First, we test our proposed method on RTTS dataset. From Table \ref{RTTS}, we can tell that compared to other 10 SOTA approaches, our D-YOLO achieves better mAP in nearly all classes. In addition, from Fig \ref{Foggy driving comparison} with the foggy driving dataset, we can still observe that our D-YOLO performs better than other candidates. This is because our specially designed feature adaption and feature fusion module allow the network to learn rich information from both hazy and normal scenarios, which is of vital importance for overcoming the influence of adverse weather.
\begin{table}[t]
\centering
\setlength{\tabcolsep}{4pt}
\caption{Quantitative comparisons on Foggy Driving Dataset.}
\label{Foggy driving comparison}
\begin{tabular}{ccccccc}
\toprule
Method              & Person & Bicycle & Car   & Motor  & Bus   & All   \\ \midrule
Yolov8            & 0.252  & \textcolor{blue}{0.208}   & 0.536 & 0.0498 & \textcolor{blue}{0.399} & 0.289 \\
Yolov8-C          & 0.249  & 0.195   & 0.517 & 0.0457 & 0.387 & 0.279 \\
IA-Yolo           & \textcolor{blue}{0.283}  & 0.185   & \textcolor{blue}{0.543} & 0.093  & 0.392 & \textcolor{blue}{0.299} \\ 
DSNet             & 0.225  & 0.203   & 0.511 & \textcolor{blue}{0.127}  & 0.334 & 0.28  \\
MS-DAYolo         & 0.256  & 0.211   & 0.498 & 0.114  & 0.369 & 0.29  \\
AOD-YOLOv8        & 0.200    & 0.160    & 0.500   & 0.146  & 0.345 & 0.27  \\
MSBDN-Yolov8      & 0.227  & 0.174   & 0.506 & 0.076  & 0.355 & 0.268 \\
Griddehaze-Yolov8 & 0.235  & 0.155   & 0.500   & 0.0514 & 0.378 & 0.264 \\
DCP-Yolov8        & 0.241  & 0.186   & 0.473 & 0.083  & 0.349 & 0.266 \\
Ours              & \textcolor{red}{0.307}  & \textcolor{red}{0.227}   & \textcolor{red}{0.576} & \textcolor{red}{0.151}  & \textcolor{red}{0.412} & \textcolor{red}{0.335} \\ \bottomrule

\end{tabular}
\end{table}

\subsubsection{Comparisons on Synthetic Dataset}
We also evaluated our model on synthetic dataset.From Table \ref{Citiscapes compparison}, results from Citiscapes-foggy dataset indicate that our D-YOLO shows greater potention at detecting objects under foggy conditions.Furthermore, we found that there is a severe drop in testing result regardless of the dehazing algorithm we use if the second stage detection model is trained on hazy images. The discrepancy between the training set(hazy images) and testing set (dehazed images) could indicate a clear domain shift, leading to reduction in detection performance. Therefore, in our experiments, the second stage detection models are always pre-trained on foggy images. 
\begin{table}[t]
\centering
\setlength{\tabcolsep}{4pt}
\caption{Quantitative comparisons on FoggyCityscapes dataset.}
\label{Citiscapes compparison}
\begin{tabular}{ccccccc}
\toprule
Method              & Person & Bicycle & Car   & Motor  & Bus   & All   \\ \midrule
Yolov8            & 0.257  & 0.159   & 0.366 & 0.042 & 0.172 & 0.199 \\
Yolov8-C          & 0.261  & 0.153   & 0.359 & 0.037 & 0.158 & 0.194 \\
IA-Yolo           & 0.267  & 0.174   & \textcolor{blue}{0.373} & 0.043 & \textcolor{blue}{0.192} & \textcolor{blue}{0.210}  \\ 
DSNet             & 0.251  & 0.143   & 0.369 & 0.045  & 0.167 & 0.195 \\
MS-DAYolo         & \textcolor{blue}{0.274}  & \textcolor{red}{0.212}   & 0.353 & 0.034 & 0.179 & 0.210 \\
AOD-YOLOv8        & 0.236  & 0.144   & 0.332 & 0.031 & 0.129 & 0.174 \\
MSBDN-Yolov8      & 0.235  & 0.127   & 0.341 & \textcolor{blue}{0.053} & 0.130  & 0.177 \\
Griddehaze-Yolov8 & 0.221  & 0.137   & 0.323 & 0.030 & 0.133 & 0.169 \\
DCP-Yolov8        & 0.243  & 0.141   & 0.339 & 0.032  & 0.161 & 0.183 \\
Ours              & \textcolor{red}{0.311}  & \textcolor{blue}{0.204}   & \textcolor{red}{0.420}  & \textcolor{red}{0.075} & \textcolor{red}{0.215} & \textcolor{red}{0.245} \\ \bottomrule
\end{tabular}
\end{table}
In IA-Yolo, image restoration and object detection are proceeded sequentially, controled by one detection loss, a self-designed DIP module is responsible for the restoration task. In DSNet, the dehazing network is AOD-net.
\subsection{Qualitive comparison}
\begin{figure*}[ht]
    \centering
    \includegraphics[width=1\textwidth]{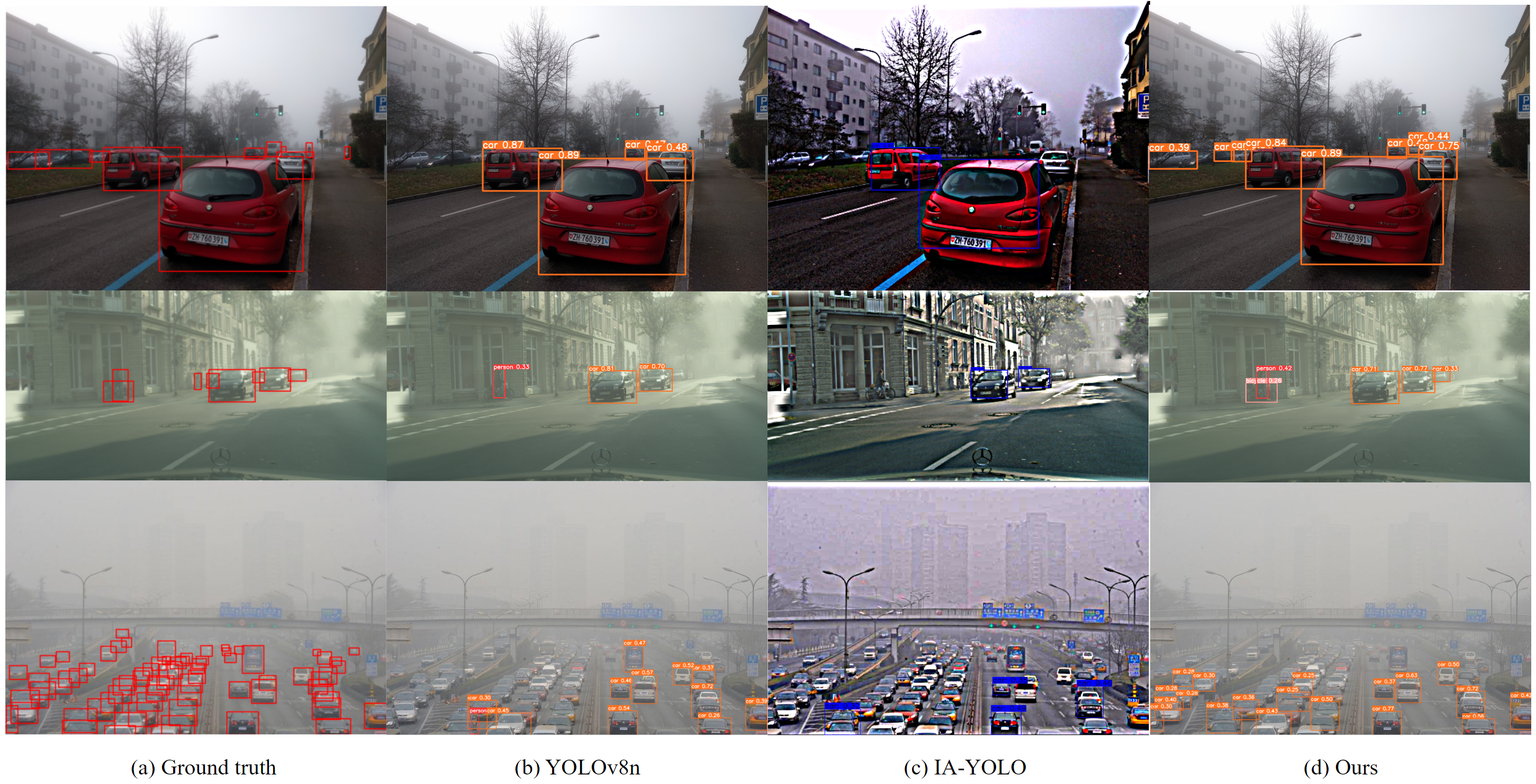}
\caption{Detection results of our proposed method. (a) Ground truth, (b) Baseline(Yolov8s), (c) our D-YOLO. As we can see, our D-YOLO generates more bounding box with higher confidence score.} 
\label{qualitive}
\end{figure*}

For qualitive comparison, we compared our D-YOLO with the SOTA method IA-Yolo, As shown in Fig \ref{qualitive}, we demonstrate Tree detection results from the FoggyCityscapes dataset. As we can see, our D-YOLO can generate more detected objects with higher accuracy and more confidence.

In IA-yolo, detection and restoration task are controled with only one loss. However, as shown in fig, The outputs of IA-Yolo look different from other models, which is because in addition to dehazing, IA-Yolo also includes a series of

traditional image processing techniques. Which is benifitial to object detection but make IA-Yolo a less robust framework for improving detection performance in hazy scenarios.
\subsection{Efficiency analysis}
We also evaluate our D-YOLO in efficiency. The inference speed and parameter amount are demonstrated in table. All experiments are conducted on RTTS dataset with a single RTX3090 GPU. Image resolution is equal to $448 \times 640 \times 3$. We compare our D-YOLO with multiple dehaze\&detect methods as well as IA-Yolo. As demonstrated in Table \ref{EXperiments speed}, D-YOLO shows great advantages in parameter amount and inference speed, ensuring real-time prediction with improved performance.

\begin{table}[t]
\centering
\caption{Efficiency analysis.}
\label{EXperiments speed}
\setlength{\tabcolsep}{12pt}
\begin{tabular}{cccc}
\toprule
Method             & Speed & FPS & mAP  \\
\midrule
Yolov8              & 0.025  & 40.0& \textcolor{blue}{0.381}     \\
AOD-YOLOv8          & 0.135 & 7.4 & 0.345\\
MSBDN-Yolov8        & 0.104 & 9.6& 0.337 \\
GridDehaze-Yolov8   & 0.071 & 14.1 & 0.371\\
DS-Net              & 0.049 & 20.4& 0.327 \\
IA-YOLO             & 0.035 & 28.6 & 0.357\\
D-YOLO              & 0.033 & 30.3& \textcolor{red}{0.430}\\
\bottomrule
\end{tabular}
\end{table}

\subsection{Experiments on rainy condition}

To further explore the generalizing ability of D-YOLO in other adverse weather conditions, we adopted RainyCityscapes dataset to evaluate the detection capability of D-YOLO in rainy conditions. RainyCityscapes dataset contains 10620 synthetic rainy images with eight annotated object classes(car, train, truck, motorbike, bus, bike, rider, person). Each corresponding clear image is paired with 36 variants. 3600 images are selected to form the trainset, 1800 images are selected as testset. As mentioned above, we only keep the aforementioned five object classes in all annotations.
\begin{figure}[t]
    \centering
    \includegraphics[width=1\linewidth]{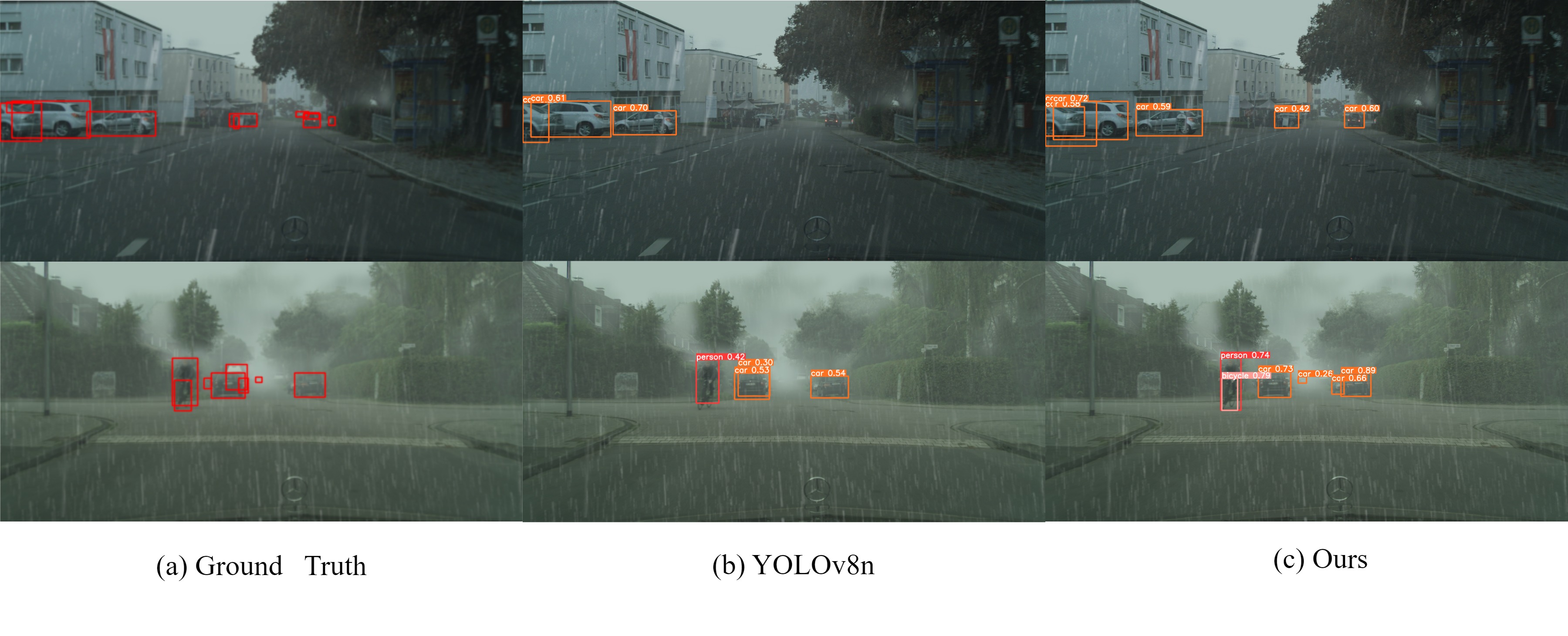}
\caption{Detection results of baseline YOLOv8n and D-YOLO on RainyCityscapes datset.(a) Ground truth, (b) baseline, (c)D-YOLO}
\label{rain}
\end{figure}
The comparsion is made among the baseline Yolov8, Yolov8-C and our D-YOLO. As shown in Table, our D-YOLO out performs other models. Fig \ref{rain} demonstrates detection results of the baseline YOLOv8n and our D-YOLO. As can be seen, our proposed network can discern more objects with higher confidence score, further approving the generalizing capability of our D-YOLO under adverse weather conditions.

\begin{table}[t]
\centering
\caption{Comparisons of baseline YOLOv8n and D-YOLO on RainyCityscapes dataset}
\label{EXperiments rainy}
\setlength{\tabcolsep}{6pt}
\begin{tabular}{@{}cccclll@{}}
\toprule
Method     & Person & Bicycle & Car   & Motor & Bus   & All   \\ \midrule
YOLOv8   & \textcolor{blue}{0.274}  & \textcolor{blue}{0.132}   & \textcolor{blue}{0.313} & \textcolor{blue}{0.049} & \textcolor{blue}{0.344} & \textcolor{blue}{0.222} \\
YOLOv8-C & 0.250  & 0.132   & 0.293 & 0.012 & 0.244 & 0.186 \\
Ours     & \textcolor{red}{0.307}  & \textcolor{red}{0.201}   & \textcolor{red}{0.404} & \textcolor{red}{0.097} & \textcolor{red}{0.428} & \textcolor{red}{0.287} \\ \bottomrule
\end{tabular}
\end{table}

\subsection{Ablation Studies and Analysis}  

Ablation studies are conducted on various module combinations, different loss functions, and different loss weights.
\subsubsection{Module combination}
In our exploration of the architecture of the D-YOLO, we systematically investigated the impact of various modules and subnetworks on the model’s overall performance. This included the clear feature extraction subnetwork, The attention feature fusion module, and the Feature Enhancement module. Notably, when omitting the attention feature fusion module, the architecture defaults to a single branch model. In addition, we also experiment on the effect of ODConv in FA module. We compare it with normal convolution and SEAttention\cite{SEAttn}, which has a similar structure with ODConv but only includes channel-wise attention weight. Our empirical results, as illustrated in Table \ref{EXperiments module combinations}, underscore the critical role that both the CFE subnetwork and the attention feature fusion module play in enhancing the performance of the entire network. The integration of these components significantly contributes to the improvement of the model’s efficacy, demonstrating their indispensability in the architectural design for optimized object detection, especially in adverse weather conditions.
\begin{table}[t]
\centering
\setlength{\tabcolsep}{4pt}
\renewcommand{\arraystretch}{1.5}
\caption{Experiments on different module combinations.}
\label{EXperiments module combinations}
\begin{tabular}{ccccccccc}
\toprule
Module      & V0 & V1 & V2 & V3 & V4 & V5 & V6 &  \\
\midrule
Dual Branch &    &    &    &    & \checkmark  & \checkmark  & \checkmark &  \\
CFE         &    & \checkmark  &    & \checkmark  & \checkmark  & \checkmark  & \checkmark  &  \\ 
FA          &    &    & \checkmark  & \checkmark  & \checkmark  & \checkmark  & \checkmark  &  \\
AFM         &    &    &    &    &    & \checkmark  & \checkmark  &  \\
ODConv      &    &    & \checkmark  & \checkmark  & \checkmark  & \checkmark  &    &  \\
mAP         & 0.381  &  0.391 & 0.383 & 0.389  & 0.417 & \textcolor{red}{0.430}  & \textcolor{blue}{0.425} &  \\
\bottomrule
\end{tabular}
\end{table}
\subsubsection{Loss function}
Furthermore, we also made comparisons among different loss functions. In our experiments, we considered five distinct variants, covering L1, L2 and KL divergence: MimicLoss, MGDLoss(matching guided distillation)\cite{MGD}, CWDLoss\cite{shu2021channel}(Channel-wise distillation) and PWDLoss. MimicLoss and MGDLoss are based on L1Loss, where as CWD and PWD are based on Kulback-Leibler(KL) divergence. As shown in Table \ref{EXperiments on loss function}, our empirical findings indicate that models with L1 based losses generally experienced a degradation in performance, which can be attributed to the stringent nature of L1 loss, which poses a strong constraint and can bring negative effect to the convergence of the model, leading to diminished model performance. Conversely, KL divergence, which serves as a measure of similarity between two probability distributions, focuses on the relative distribution of features and tends to mitigate the influence of irrelevant background information. Networks trained under the constraint of KL divergence exhibited enhanced robustness to adverse weather conditions and a significant improvement in accuracy. Among the evaluated losses, CWDLoss demonstrates superior performance. This is because there exists a domain shift between foggy images and clear images, primarily manifesting as differences in the channel dimensions. Narrowing the distance in the channel distributions between $F_d$ and $F_c$ helps to enhance the model’s performance in foggy conditions.  

\begin{table}[t]
\centering
\setlength{\tabcolsep}{20pt}
\caption{EXperiments on loss function.}
\label{EXperiments on loss function}
\begin{tabular}{ccc}
\toprule
Type   & Name  & mAP   \\
\midrule
L1     & Mimic-L1 & 0.347 \\
L1     & MGD   & 0.372 \\
L2     & Mimic-L2 & 0.347 \\
KL div & CWD   & \textcolor{red}{0.430}  \\
KL div & PWD   & \textcolor{blue}{0.398}\\
\bottomrule
\end{tabular}
\end{table}

\subsubsection{Attention mechanism}
\begin{figure}[t]
    \centering
    \includegraphics[width=1\linewidth]{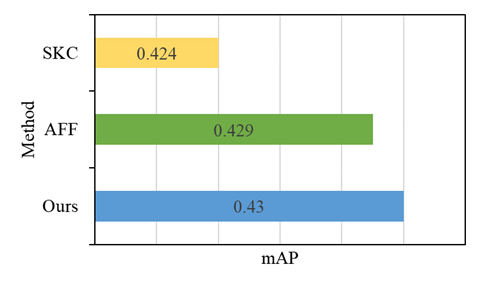}
\caption{Comparisons on different attention mechanisms in our attention feature fusion module.}
\label{fig:AFF experiments}
\end{figure}

We study the effect of different attention mechanisms within our Attention feature fusion module, SKC(selective kernel convolution), AFF(attentional feature fusion) are chosen as competitors in our comparisons. In SKC, channel attention is applied through global average pooling, followed by a fully connected layer and a softmax function. In AFF, attention is calculated from a dual-branch structure, extracting global and local information separately. Our method adopts $r \times r$ average pooling with a shortcut structure, and replaces the fully connected layer with a convolution layer, reducing complexity while maintaining a single branch structure. This design allows our attention feature fusion module to process both channel and location information at the same time with low computational cost. We compared different attention mechanisms on RTTS dataset, results are shown in Fig \ref{fig:AFF experiments}, our attention feature fusion model outperforms other competitors in mAP.
\subsection{Loss weight}
To further improve the detection performance of D-YOLO under adverse weather, several losses were applied. The loss in our study comprises two main parts: Detection loss $L_d$ and Dehazing Loss $L_r$. Correspondingly, $\lambda_1$ and $\lambda_2$ are utilized to adjust the proportions of $L_r$ and $L_d$. To explore the best combination of $L_r$ and $L_d$, we conducted extensive experiments on RTTS dataset. To, be noticed, beyond fixed weight parameters, we employed a dynamic weight with gradient penalty. Throughout an epoch, as training progresses, the proportion of $\lambda_r$ gradually decreases to 1. As evidenced by the data from Table \ref{EXperiments loss weight}, the introduction of $L_r$ significantly contributes to enhancing model performance, with the most effective results being observed when $L_r$ is dynamically weighted.
\begin{table}[t]
\centering
\caption{EXperiments loss weight.}
\label{EXperiments loss weight}
\setlength{\tabcolsep}{4pt}
\begin{tabular}{ccccccc}
\toprule
Loss & V0 & V1  & V2  & V3  & V4  & V5 \\
\midrule
$\lambda_1$   & 1  & 0.2 & 0.3 & 0.5 & 0.2 & 1  \\
$\lambda_2$   & 1  & 0.8 & 0.7 & 0.5 & 1   & Dy \\
mAP         & 0.419 & \textcolor{blue}{0.421} & 0.417 & 0.413 & 0.409   & \textcolor{red}{0.430}\\
\bottomrule
\end{tabular}
\end{table}


\section{Conclusion}\label{sec:con} 
In this paper, we propose a unified attention framework for object detection in adverse weather conditions, called D-YOLO. There are three key components in our D-YOLO. Clear Feature extraction subnetwork is responsible for extracting haze-free features from undegraded images. The output from clear features extraction module are sent to feature adaption module where dehazed features are generated via domain adaption. In addition, we also develop an attention feature fusion module which fully integrates both hazy and dehazed features, effectively improving the complementarity and richness of target features. Qualitative and quantitative evaluations on both synthetic and real-world dataset demonstrate that D-YOLO is superior to existing state-of-the-art algorithms. However, D-YOLO still have difficulties at discerning objects in rather challenging scenarios. In the future, it will be a valuable research direction by integrating transfer learning and developing more effective feature adaption methods.

\bibliographystyle{IEEEtran}
\bibliography{ref} 




\end{document}